\documentclass[cameraready]{Interspeech}
\usepackage{cite}
\usepackage{amsmath,amssymb,amsfonts}
\usepackage{amsmath}
\usepackage{algorithmic}
\usepackage{graphicx}
\usepackage{textcomp}
\usepackage{xcolor}
\usepackage{multirow,tabularx,makecell,array}
\usepackage{hyperref} 
\usepackage{tabularx}
\usepackage{bm}        
\def\BibTeX{{\rm B\kern-.05em{\sc i\kern-.025em b}\kern-.08em
    T\kern-.1667em\lower.7ex\hbox{E}\kern-.125emX}}
\usepackage{tabularx}
\newcommand{\blfootnote}[1]{%
    \begingroup
    \renewcommand\thefootnote{}
    \footnote{#1}%
    \endgroup}
\makeatother
\newcolumntype{Y}{>{\centering\arraybackslash}X}

\title{EmoInstruct-TTS: Dual-Path Instruction-Guided Emotional Speech Synthesis}

\author[affiliation={1,2}]{Minghui}{Wu}
\author[affiliation={1,2^{*}}]{Ganjun}{Liu}
\author[affiliation={1}]{Zikun}{Fang}
\author[affiliation={2}]{Ting}{Meng}
\author[affiliation={2}]{Hongchuan}{Wu}
\author[affiliation={2}]{Bingao}{Xu} 
\author[affiliation={2}]{Yonglong}{Cai}
\author[affiliation={3}]{Jiasheng}{Chen}
\author[affiliation={1}]{Jun}{Du}

\address{
  $^1$ University of Science and Technology of China, China\\
  $^2$ iFLYTEK Research, China\\
  $^3$ Huawei Technologies Co., Ltd., China
}

\email{mhwu@iflytek.com, gjliu4@iflytek.com, fangzk23@mail.ustc.edu.cn, tingmeng@iflytek.com, hcwu4@iflytek.com, baxu@iflytek.com, ylcai9@iflytek.com, chenjiasheng@huawei.com, jundu@ustc.edu.cn}


\usepackage{comment}

\begin{document}

\maketitle

\begin{abstract}
\blfootnote{* Indicates corresponding authors.}
    Instruction-based controllable speech synthesis enables users to specify emotions through natural language. However, existing approaches often rely on coarse emotion labels and lack explicit modeling of fine-grained intensity. We propose EmoInstruct-TTS, a dual-path instruction-guided framework for emotional speech synthesis. We introduce Emotion2embed, a supervised semantic-acoustic emotion embedding covering 48 emotional states, including fine-grained categories and intensity levels. To infer embeddings from free-form instructions, we design an Instruction-Conditioned Emotion Flow Model (ICE-Flow) that generates acoustically grounded emotion representations. The inferred embeddings are integrated into an LLM-based synthesis pipeline to provide explicit emotional control while preserving semantic planning. Experiments show improved emotional controllability and speech naturalness over strong baselines.\footnotetext{Audio demos are available at \textit{https://huanyu-lab.github.io/EMOINSTRUCT-TTS}}
\end{abstract}

\section{Introduction}
\label{sec:intro}


Recent advances in neural text-to-speech (TTS) models have achieved speech naturalness and intelligibility approaching human perception~\cite{NEURIPS2021_748d6b6e, kim2022naturalspeech, shen2023naturalspeech2, xie2025fireredtts, chen2025f5, jiang2025megatts}. As TTS technology matures, research has increasingly shifted from neutral speech generation to controllable synthesis, where users specify attributes such as emotion, style, and prosody through explicit control signals~\cite{wang2023styletts2, gao2022valle, tan2024emotional, EME-TTS2025, yamamoto2025description}. Among these attributes, emotional expressiveness plays a key role in enhancing engagement and personalization in applications such as virtual assistants, audiobooks, and conversational agents~\cite{kreuk2023audiolm, parlerTTS2023, MPE-TTS2025, qiang2025instructaudio}.

\begin{figure}[t]

\begin{minipage}[b]{.49\linewidth}
  \centering
  \centerline{\includegraphics[width=\linewidth,trim=10 8 10 8,clip]{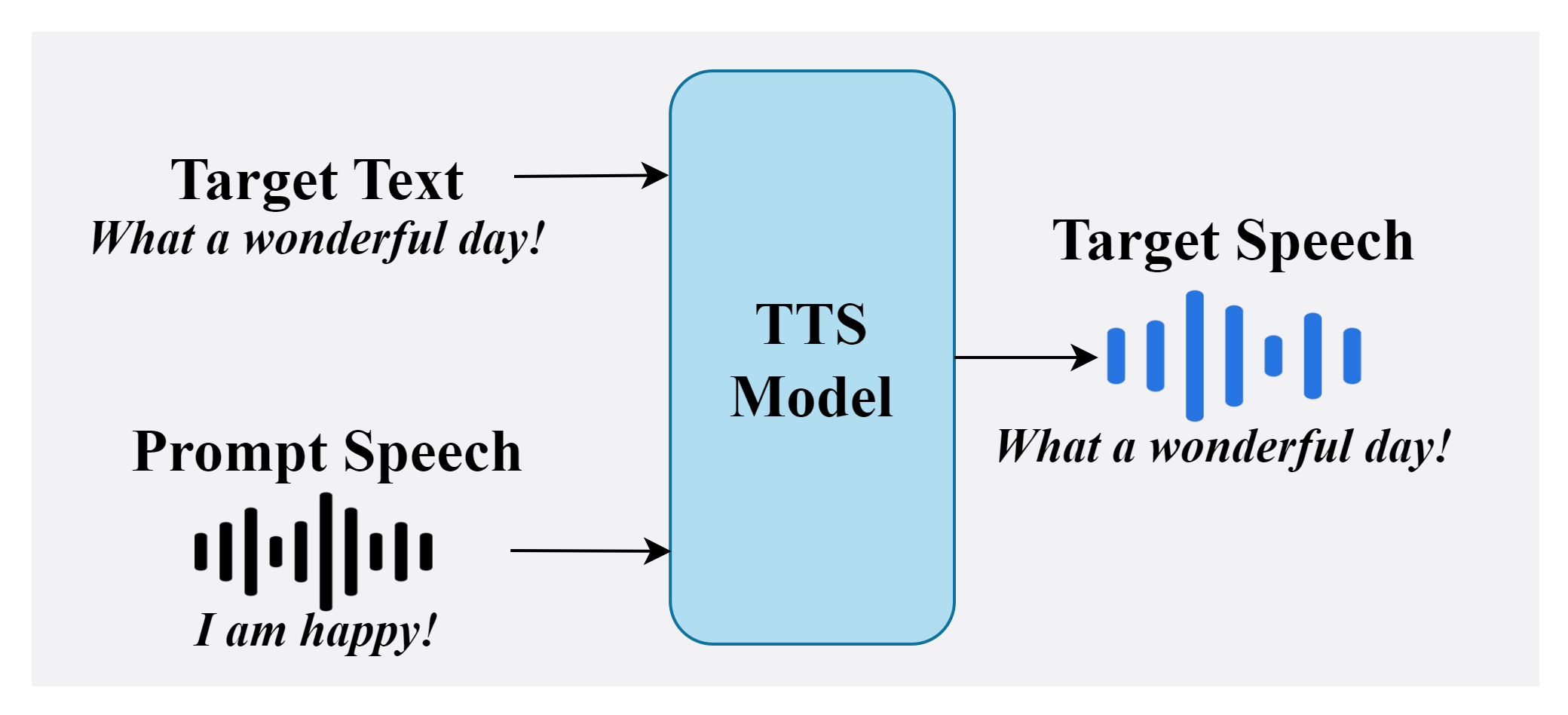}}
  \centerline{(a) Prompt speech-driven}\medskip
\end{minipage}
\hfill
\begin{minipage}[b]{.50\linewidth}
  \centering
  \centerline{\includegraphics[width=\linewidth,trim=10 8 10 8,clip]{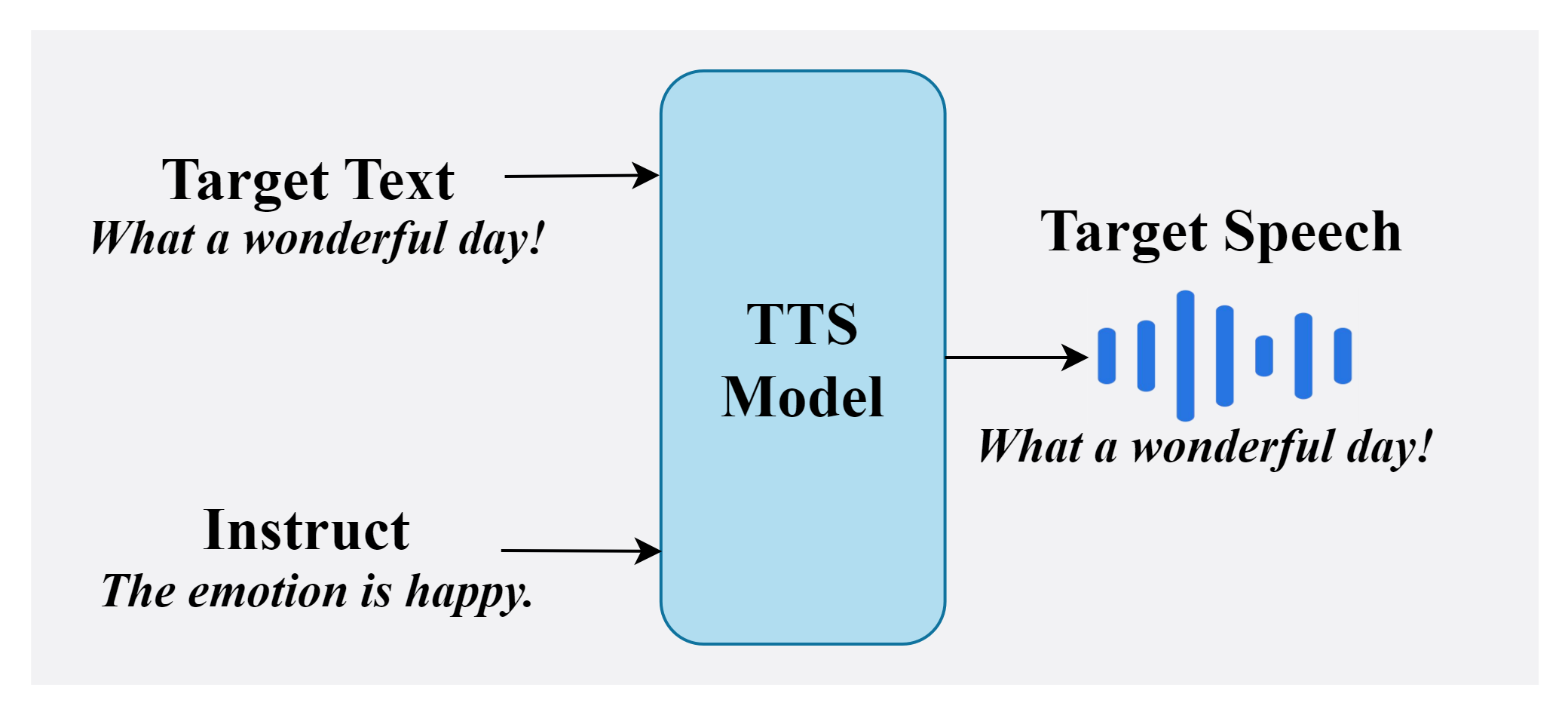}}
  \centerline{(b) Text instruction-driven}\medskip
\end{minipage}
\begin{minipage}[b]{1.0\linewidth}
  \centering
  \centerline{\includegraphics[width=8.5cm]{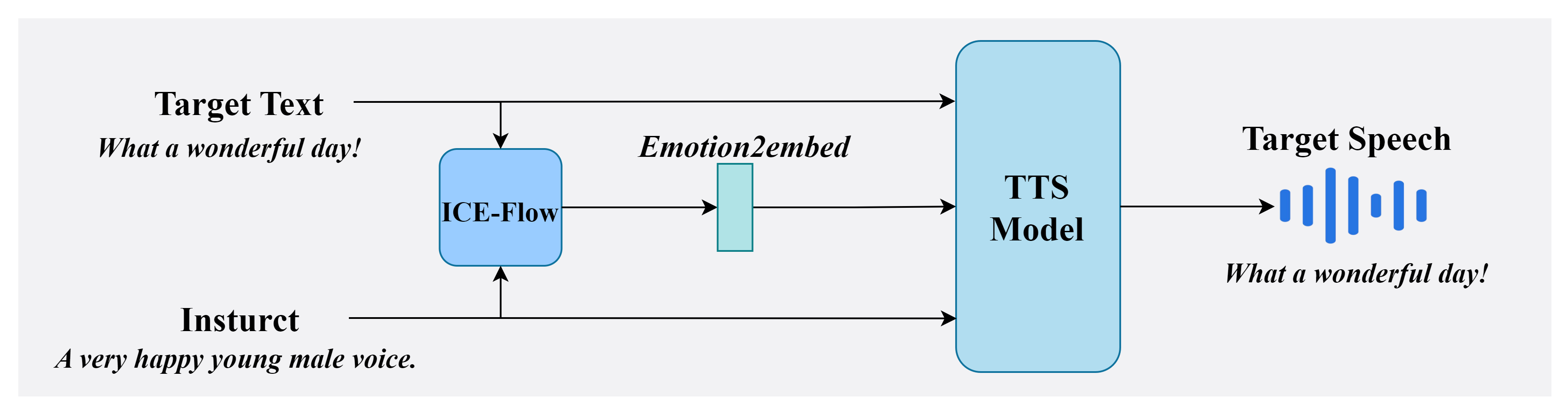}}
  \centerline{(c) Proposed framework}\medskip
\end{minipage}
\caption{Comparison between previous and proposed emotional speech synthesis frameworks.}
\label{fig:compar}

\end{figure}
\begin{figure*}[t]
    \centering
    \includegraphics[width=.9\linewidth]{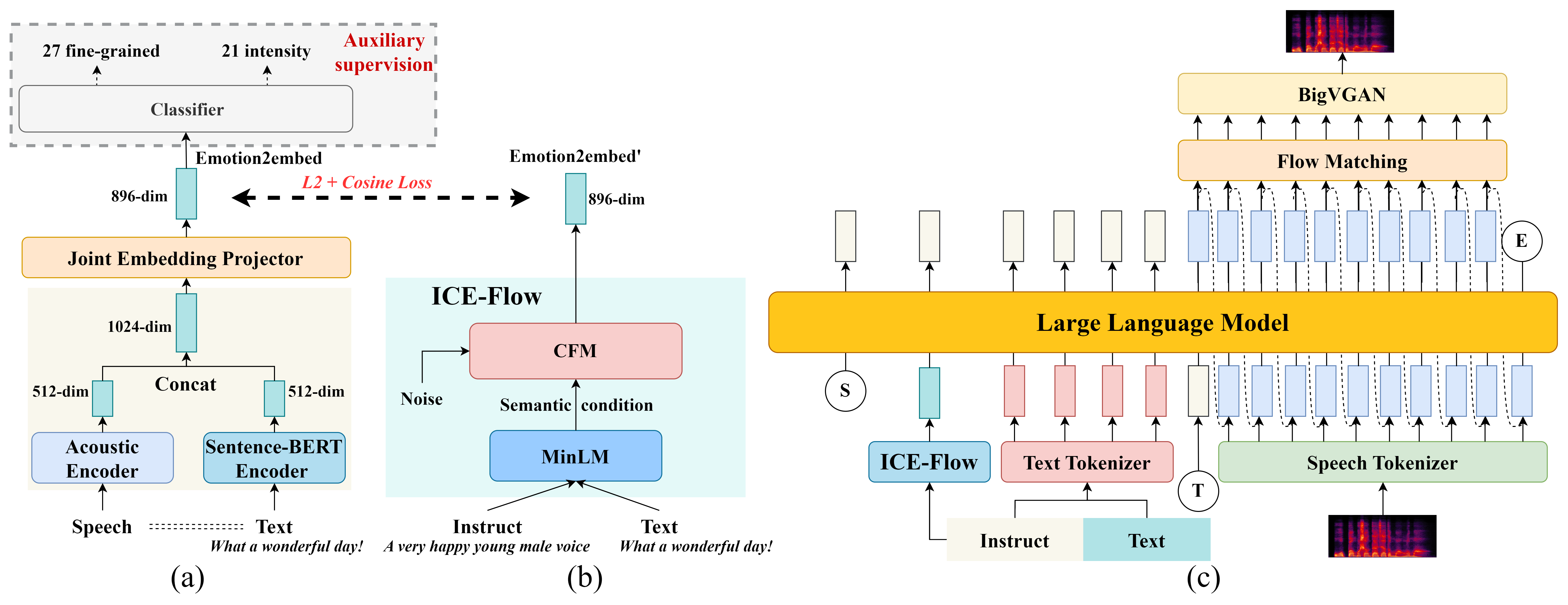}
    \caption{The dual-path instruction-guided synthesis pipeline of EmoInstruct-TTS. (a) Training process for acquiring the Emotion2embed representation. (b) Generation of Emotion2embed via the ICE-Flow model. (c) Overview of the EmoInstruct-TTS synthesis system.}
    \label{fig:framework}
\end{figure*}
Existing emotional TTS approaches mainly follow two paradigms. 
The first conditions synthesis on reference emotional speech or latent style embeddings extracted from emotional recordings~\cite{min2023mintts, chen2024emoKnob, SA-RAS2025, ju2024naturalspeech, zhou2025indextts2} (Fig.~\ref{fig:compar}(a)). 
While effective, these methods typically require curated reference data and often suffer from speaker dependency and timbre mismatch. 

The second paradigm focuses on instruction-driven controllability, where natural language descriptions guide speech generation through large language models or instruction prompts~\cite{parlerTTS2023, du2024CosyVoice, du2024CosyVoice2, du2025CosyVoice3, hu2026qwen3, hu2026voicesculptor} (Fig.~\ref{fig:compar}(b)). 
Although flexible, existing instruction-based systems usually rely on coarse emotion categories and lack explicit mechanisms to model fine-grained emotional variation and intensity, resulting in unstable emotional control~\cite{emopro, emoVoice2025}. 
Recent studies further suggest that linguistic instructions alone are often insufficient to capture detailed acoustic correlates of emotion~\cite{minmax2023}.

To address these challenges, we propose \textit{EmoInstruct-TTS}, a \textbf{dual-path instruction-guided framework} for controllable emotional speech synthesis. The key idea is to combine natural language instructions for semantic guidance with structured emotion embeddings for fine-grained emotional modulation.

Specifically, we introduce \textbf{Emotion2embed}, a semantic--acoustic emotion representation covering 48 emotional states, including 27 fine-grained emotion categories and 21 emotion--intensity combinations derived from seven primary emotions with three intensity levels. To infer emotion embeddings from free-form instructions, we further propose an \textbf{Instruction-Conditioned Emotion Flow Model (ICE-Flow)} that generates instruction-conditioned emotion embeddings from natural language descriptions. 

The inferred emotion embeddings are integrated into an LLM-based synthesis pipeline together with instruction text, enabling complementary roles of semantic planning and explicit emotional control (Fig.~\ref{fig:compar}(c)).

Experimental results show that EmoInstruct-TTS improves emotional controllability and speech naturalness over strong emotional and instruction-driven TTS baselines, particularly in modeling fine-grained emotion intensity variations.

The main contributions of this work are threefold:
\begin{itemize}
\item We propose \textbf{Emotion2embed}, a structured semantic--acoustic emotion representation for modeling fine-grained emotion categories and intensity variations.
\item We introduce \textbf{ICE-Flow}, an instruction-conditioned model that generates acoustically grounded emotion embeddings from natural language instructions.
\item We present a \textbf{dual-path instruction-guided TTS framework} that combines semantic instruction processing with explicit emotion embeddings for controllable emotional speech synthesis.
\end{itemize}

\section{Method}
\label{sec:method}
EmoInstruct-TTS is a disentangled framework for instruction-guided emotional speech synthesis. As shown in Fig.~\ref{fig:framework}, it consists of three components: an instruction-conditioned emotion generator, an LLM-based semantic encoder, and a speaker-conditioned TTS decoder followed by a neural vocoder. The framework separates semantic planning from emotion-specific acoustic control.

\subsection{Instruction-Conditioned Emotion Generator}
\label{ssec:CLAP}
\subsubsection{\textbf{Emotion2embed: Structured Semantic--Acoustic Emotion Representation}}
\label{sssec:emotion2embed}


We introduce \textit{Emotion2embed}, a semantic-acoustic embedding for fine-grained emotional modeling that encodes both emotion category and ordered intensity structure while remaining speaker-invariant.

\textbf{Semantic-acoustic encoding.}
Given emotional speech $x$ and its paired text description $t$, we extract complementary semantic and acoustic features using a Sentence-BERT encoder (bge-large-zh v1.5)~\cite{bge_embedding} and an ECAPA-TDNN encoder~\cite{desplanques2020ecapa}, respectively. The two features are concatenated and projected into a unified embedding space:




\begin{equation}
\mathbf{z}_{\text{emo}} = W [f_{\text{text}}(t); f_{\text{acoustic}}(x)] + b,
\end{equation}
where $\mathbf{z}_{\text{emo}} \in \mathbb{R}^{896}$ in our implementation.

\textbf{Supervised emotion and intensity structuring}
Emotion2embed is trained using a multi-task objective combining emotion classification, intensity classification, and an ordinal intensity constraint:

\begin{equation}
\mathcal{L}_{\text{emo}} =
\mathcal{L}_{\text{cls}}^{\text{emo}}
+ \lambda \mathcal{L}_{\text{cls}}^{\text{int}}
+ \beta \mathcal{L}_{\text{ord}}.
\end{equation}

\textbf{Ordinal intensity geometry}
To enforce intensity ordering, embeddings are $\ell_2$-normalized and associated with a learnable direction vector $\mathbf{u}_e$. A margin-based ranking loss enforces monotonic ordering of low, medium, and high intensity levels:

\begin{equation}
\mathcal{L}_{\text{ord}} =
\max(0, m - (s_{\text{mid}} - s_{\text{low}}))
+ \max(0, m - (s_{\text{high}} - s_{\text{mid}})).
\end{equation}


\subsubsection{\textbf{Instruction-to-Emotion2embed Mapping via ICE-Flow}}
\label{sssec:Emotion2embedMapping}


We propose \emph{ICE-Flow} to generate emotion embeddings from natural language instructions, mapping instruction semantics to Emotion2embed representations under acoustic supervision.

Given an instruction text $t$, a multilingual MiniLM encoder encodes semantic features $\mathbf{h}_{\text{text}}$~\cite{allMiniLMv2}. 
ICE-Flow models a conditional distribution over emotion embeddings:

\begin{equation}
p(\mathbf{z}_{\text{emo}} \mid \mathbf{h}_{\text{text}}).
\end{equation}
where $\mathbf{z}_{\text{emo}}$ denotes the target Emotion2embed representation.

\textbf{Sample-level acoustic grounding}
During training, ICE-Flow is supervised using sample-level Emotion2embed targets extracted from real emotional speech:
\begin{equation}
\mathbf{z}_{\text{emo}} = f_{\text{Emotion2embed}}(x, t).
\end{equation}
The model is optimized via a regression objective:

\begin{equation}
\mathcal{L}_{\text{sample}} =
\mathbb{E}_{(t,x)}
\bigl\lVert
f_{\text{ICE}}(\mathbf{h}_{\text{text}}(t)) - \mathbf{z}_{\text{emo}}(x)
\bigr\rVert_2^2,
\end{equation}

ensuring that instruction-generated embeddings reflect variability induced by real acoustic realizations.

\textbf{Distribution-level regularization}
To discourage mode collapse and preserve realistic variation, we further align the covariance of sampled embeddings with that of real emotion embeddings sharing the same emotion--intensity label:
\begin{equation}
\mathcal{L}_{\text{dist}} =
\bigl\lVert
\text{Cov}(\hat{\mathbf{z}}_{\text{emo}}) -
\text{Cov}(\mathbf{z}_{\text{emo}}^{\text{real}})
\bigr\rVert_F.
\end{equation}

The final objective is
\begin{equation}
\mathcal{L}_{\text{ICE}} =
\mathcal{L}_{\text{sample}} + \gamma \mathcal{L}_{\text{dist}}.
\end{equation}

At inference, ICE-Flow generates emotion embeddings conditioned solely on instruction semantics.
Classifier-free guidance is applied to control the strength of instruction adherence.

\subsection{Dual-Path Instruction-Guided Speech Synthesis}
\label{ssec:emoinstruct-tts}





We adopt a dual-path design that separates semantic modeling from emotion-specific acoustic control. Instruction information is decomposed into two components: (i) semantic guidance for linguistic realization and (ii) explicit emotion embeddings for acoustic modulation.

\noindent \textbf{Instruction-to-Emotion2embed Path.}
This path generates emotion control signals from natural language instructions. The instruction is encoded by a MiniLM encoder and mapped to an emotion embedding via ICE-Flow:

\begin{equation}
\label{eq:emotion_embed}
\mathbf{E}_{\text{emo}} = f_{\text{ICE-Flow}}(f_{\text{MiniLM}}(\text{instruction})),
\end{equation}
where $\mathbf{E}_{\text{emo}} \in \mathbb{R}^{d}$ denotes the generated Emotion2embed.


\noindent \textbf{Instruction-to-LLM Path.}
A LLM processes the instruction and target text to generate semantic tokens representing linguistic content and contextual intent:

\begin{equation}
\label{eq:semantic}
\mathbf{S} = f_{\text{LLM}}(\text{instruction}, \text{text}, \mathbf{E}_{\text{emo}}),
\end{equation}
where $\mathbf{S}$ denotes the semantic token sequence produced by the LLM.



\noindent \textbf{Speech Generation.}
Semantic tokens $\mathbf{S}$, emotion embedding $\mathbf{E}_{\text{emo}}$, and speaker embedding $\mathbf{E}_{\text{spk}}$ are jointly conditioned to generate a mel-spectrogram:

\begin{equation}
\label{eq:mel}
\mathbf{M} = f_{\text{CFM\_TTS}}(\mathbf{S}, \mathbf{E}_{\text{emo}}, \mathbf{E}_{\text{spk}}),
\end{equation}
where $\mathbf{M}$ denotes the synthesized mel-spectrogram.
A high-fidelity neural vocoder (BigVGAN) then converts $\mathbf{M}$ into the output waveform:
\begin{equation}
\label{eq:vocoder}
\hat{\mathbf{x}} = f_{\text{BigVGAN}}(\mathbf{M}).
\end{equation}


\begin{figure*}[htb]

\begin{minipage}[b]{.25\linewidth}
  \centering
  \centerline{\includegraphics[width=6cm]{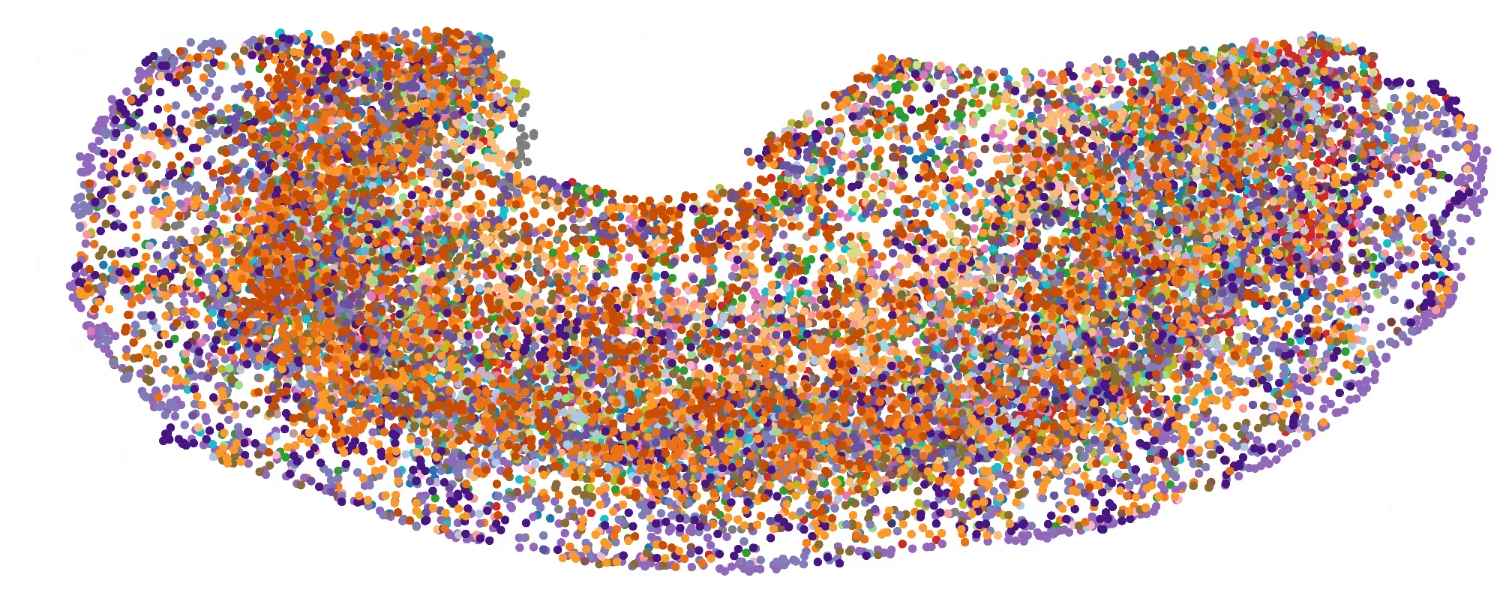}}
  \centerline{(a) Emotion2vec: 27 emotions}\medskip
\end{minipage}
\hfill
\begin{minipage}[b]{0.25\linewidth}
  \centering
  \centerline{\includegraphics[width=6cm]{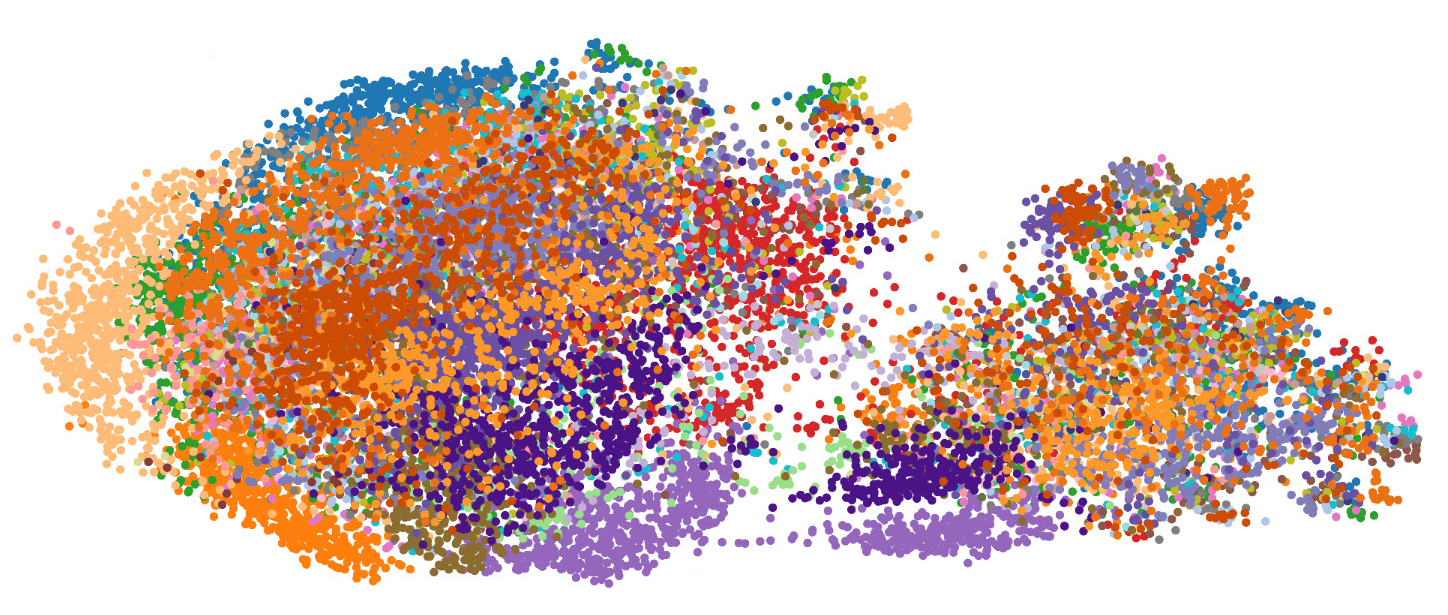}}
  \centerline{(b) Emotion2embed: 27 emotions}\medskip
\end{minipage}
\hfill
\begin{minipage}[b]{.25\linewidth}
  \centering
  \centerline{\includegraphics[width=6cm]{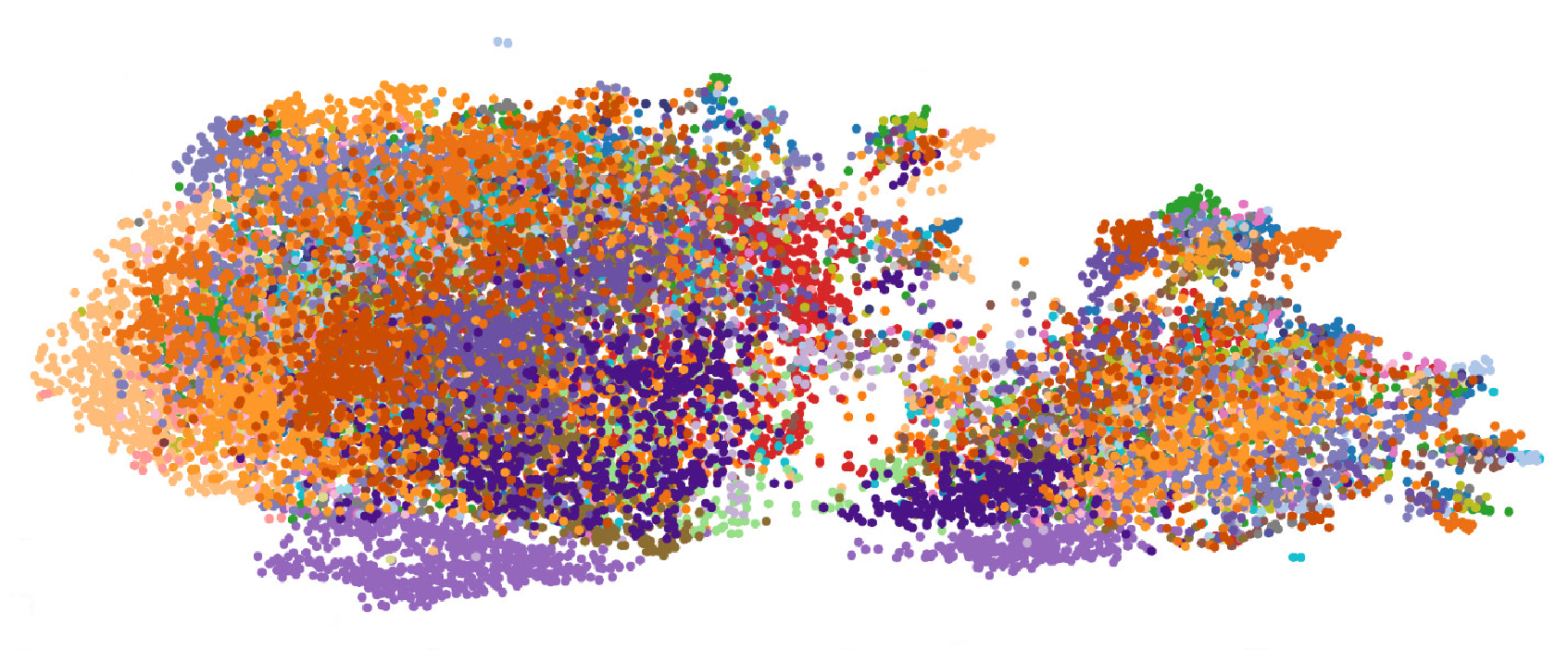}}
  \centerline{(c) ICE-Flow Emotion2embed: 27 emotions}\medskip
\end{minipage}
\hfill
\begin{minipage}[b]{.25\linewidth}
  \centering
  \centerline{\includegraphics[width=6cm]{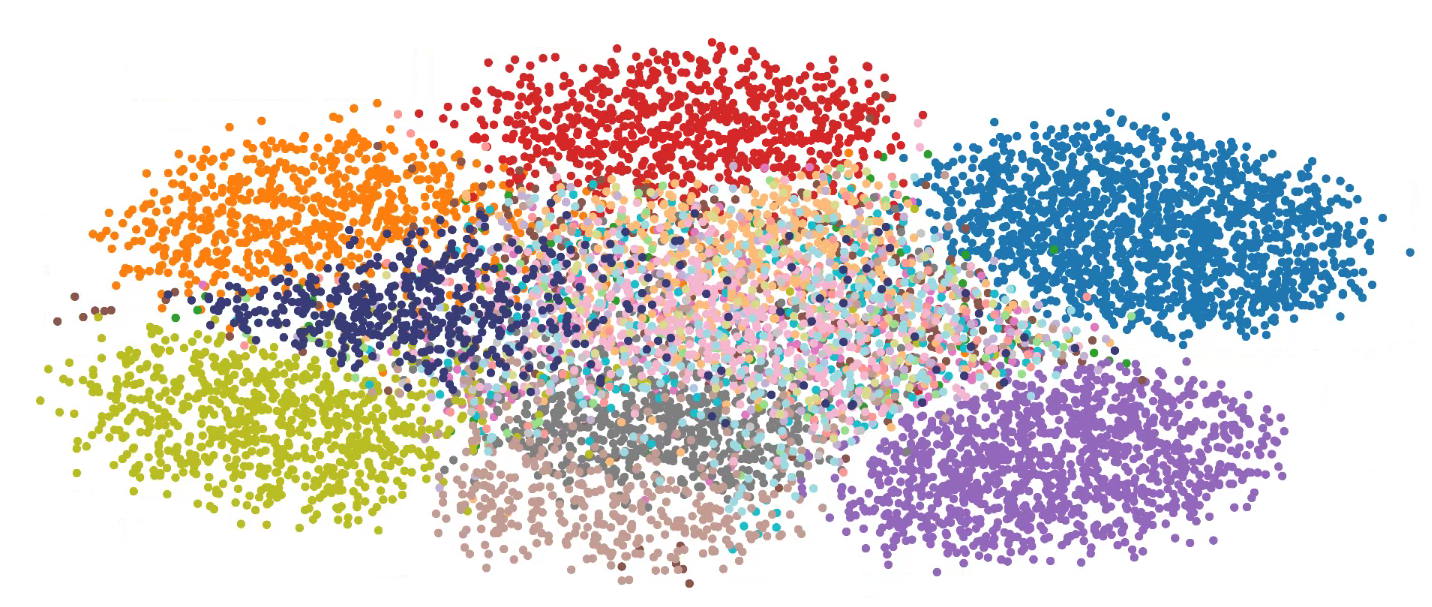}}
  \centerline{(d) Emotion2vec: 21 intensity levels}\medskip
\end{minipage}
\hfill
\begin{minipage}[b]{.25\linewidth}
  \centering
  \centerline{\includegraphics[width=6cm]{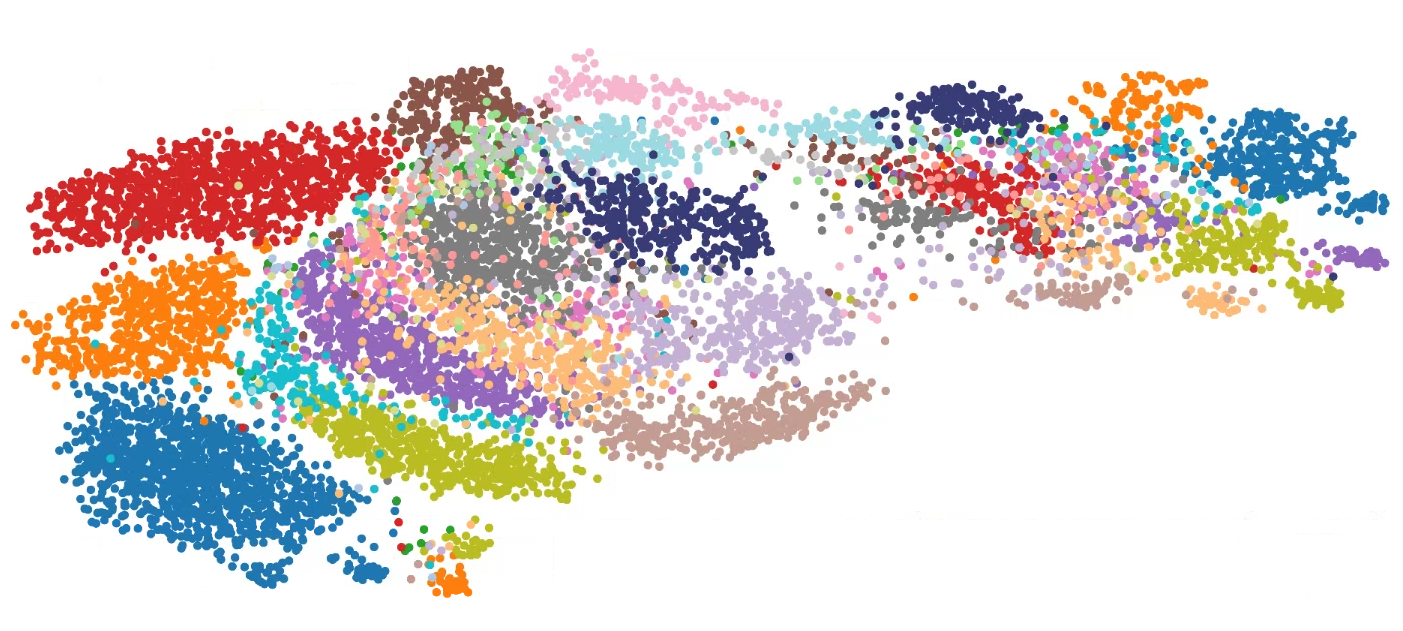}}
  \centerline{(e) Emotion2embed: 21 intensity levels}\medskip
\end{minipage}
\hfill
\begin{minipage}[b]{.25\linewidth}
  \centering
  \centerline{\includegraphics[width=6cm]{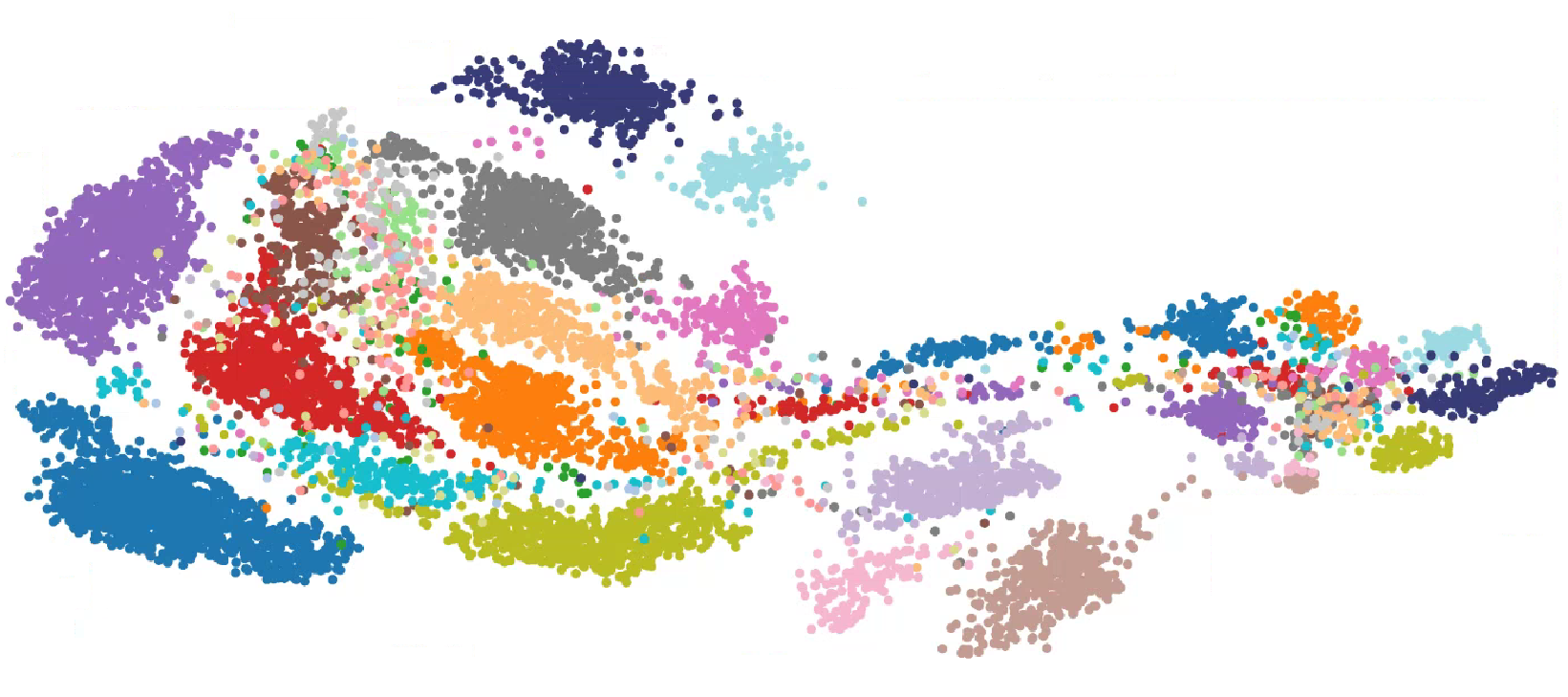}}
  \centerline{(f) ICE-Flow Emotion2embed: 21 intensity levels}\medskip
\end{minipage}
\caption{The representation performance of emoition2vec, emoition2embed, and ICE-Flow emoition2embed.}
\label{fig:perf}

\end{figure*}

\section{Experiments}
\label{sec:Exper}
\subsection{Datasets and Training Setup}
\label{ssec:dataset}
We use the Emotional Speech Dataset (ESD)~\cite{zhou2021seen} and the Chinese Natural Complex Emotion Dataset (CNCED)~\cite{wu2025cnsced}. 
The training data consist of two subsets constructed through a semi-automatic annotation pipeline.

\textit{Dataset-Base} contains 49,903 utterances with automatically generated emotion captions produced by Gemini-2.5 Pro~\cite{comanici2025gemini} using speech audio and dataset metadata (e.g., gender, age, emotion). 
These captions provide weak semantic supervision for large-scale pretraining.

\textit{Dataset-Annotation} includes 28,402 utterances with manually verified emotion labels. 
High-expressiveness samples are selected from the base corpus and annotated with fine-grained emotion categories and intensity levels. 
The final label set covers 27 emotion categories and three intensity levels (low, medium, high) for seven primary emotions, resulting in 21 emotion–intensity combinations. 
A total of 7,600 utterances are reserved for evaluation.

ICE-Flow is trained on \textit{Dataset-Base} for 100 epochs (batch size 2048) using Adam with cosine annealing (initial learning rate $1\times10^{-4}$) and then fine-tuned on \textit{Dataset-Annotation}. 
The LLM backbone (Qwen2.5-0.5B) is adapted with LoRA (rank 32, 9.87M parameters) conditioned on instruction text, content text, and Emotion2embed. 
Acoustic features are generated using CFM and synthesized with a BigVGAN vocoder~\cite{BigVGAN}. 
Speaker embeddings and reference speech preserve speaker identity. 
\textit{CosyVoice2} and \textit{CosyVoice3} are used as baselines.
\subsection{Emotion Representation Analysis}
\label{ssec:Representation}
Figure~\ref{fig:perf} compares embeddings produced by Emotion2vec~\cite{emotion2vec}, Emotion2embed, and ICE-Flow-generated Emotion2embed across 27 emotion categories and 21 emotion--intensity levels.

\textbf{t-SNE visualization.}
Emotion2vec (Fig.~\ref{fig:perf}(a,d)) exhibits strong category overlap and weak intensity organization. 
Emotion2embed (Fig.~\ref{fig:perf}(b,e)) forms more compact clusters with consistent intensity trends within each primary emotion. 
ICE-Flow-generated embeddings (Fig.~\ref{fig:perf}(c,f)) show similar structures with smoother category boundaries while preserving intensity ordering.

\textbf{Distribution consistency.}
Table~\ref{tab:iceflow_dist} evaluates whether generated embeddings collapse to class prototypes or capture realistic variability. 
Compared variants include \textbf{ICE-Mean}, \textbf{ICE-Sample}, and \textbf{ICE-Sample+Dist}. 
Prototype regression in ICE-Mean (PCS=0.89, VR=0.65) is reduced by sample-level supervision (PCS=0.73, VR=0.88) and further improved with distribution regularization (PCS=0.67, VR=0.96, SWD=0.22). 
Intensity ordering accuracy also increases from 0.78 to 0.91.

\begin{table}[t]
\centering
\caption{Distributional consistency of instruction-generated emotion embeddings.
PCS: Prototype Collapse Score ($\downarrow$), VR: Variance Ratio (closer to $1$ is better), SWD: Sliced Wasserstein Distance ($\downarrow$), IOA: Intensity Ordering Accuracy ($\uparrow$).
}
\label{tab:iceflow_dist}
\vspace{0.5em}
\setlength{\tabcolsep}{6pt}
\begin{tabular}{l c c c c }
\hline
\textbf{Variant} & \textbf{PCS} $\downarrow$ & \textbf{VR} $\approx 1$ & \textbf{SWD} $\downarrow$ & \textbf{IOA} $\uparrow$ \\
\hline
ICE-Mean & 0.89 & 0.65 & 0.58 & 0.78 \\
\hline
ICE-Sample & 0.73 & 0.88 & 0.35 & 0.86 \\
\hline
ICE-Sample+Dist & 0.67 & 0.96 & 0.22 & 0.91 \\
\hline
\end{tabular}
\end{table}


\begin{table}[t]
\centering
\caption{Comparison and ablation results on 21 emotion–intensity speech synthesis tasks (male/female, zero-shot).
MOS, ESMOS, and SSMOS are reported. Best results are highlighted in \textbf{bold}.}
\vspace{0.5em}
\label{tab:emo-tts21}
\scriptsize
\setlength{\tabcolsep}{2.3pt}    
\renewcommand{\arraystretch}{1.15}
\begin{tabularx}{\columnwidth}{ l  c  *{3}{Y }}
\hline
Speaker & Model & MOS~$\uparrow$ & ESMOS~$\uparrow$ & SSMOS~$\uparrow$ \\
\hline
\multirow{10}{*}{Female}
 & CosyVoice2   & $4.18 \pm 0.09$ & $3.92 \pm 0.14$ & $4.46 \pm 0.10$ \\
\cline{2-5}
& CosyVoice3        & $4.15 \pm 0.10$ & $3.98 \pm 0.15$ & $4.47 \pm 0.09$ \\
\cline{2-5}
 & \makecell[c]{EmoInstruct-TTS\\(Dual-Path)}& {\boldmath $4.28 \pm 0.08$} & {\boldmath $4.25 \pm 0.12$} & {\boldmath $4.52 \pm 0.09$} \\
\cline{2-5} 
 & \makecell[c]{-w/o Emo2emb\\(Text-Only)}
   & $4.12 \pm 0.10$ & $3.78 \pm 0.16$ & $4.40 \pm 0.10$ \\
\cline{2-5} 
 & \makecell[c]{-w/o Text Instruct\\(Emo2emb-Only)}
   & $4.16 \pm 0.11$ & $3.99 \pm 0.14$ & $4.44 \pm 0.11$ \\
\hline
\multirow{10}{*}{Male}
 & CosyVoice2   & $4.14 \pm 0.10$ & $3.80 \pm 0.16$ & $4.50 \pm 0.09$ \\
\cline{2-5}
& CosyVoice3  & $4.08 \pm 0.11$ & $3.92 \pm 0.15$ & {\boldmath $4.55 \pm 0.08$} \\
\cline{2-5}
 & \makecell[c]{EmoInstruct-TTS\\(Dual-Path)} & {\boldmath $4.25 \pm 0.09$} & {\boldmath $4.10 \pm 0.13$} & $4.48 \pm 0.10$ \\
\cline{2-5} 
 & \makecell[c]{-w/o Emo2emb\\(Text-Only)}
   & $4.05 \pm 0.11$ & $3.70 \pm 0.17$ & $4.42 \pm 0.11$ \\
\cline{2-5} 
 & \makecell[c]{-w/o Text Instruct\\(Emo2emb-Only)}
   & $4.09 \pm 0.10$ & $3.86 \pm 0.16$ & $4.44 \pm 0.10$ \\
\hline
\end{tabularx}
\end{table}

\begin{table}[t]
\centering
\caption{Comparison and ablation results on 27 fine-grained emotional speech synthesis tasks (male/female, zero-shot).
MOS, ESMOS, and SSMOS are reported. Best results are highlighted in \textbf{bold}.}
\vspace{0.5em}
\label{tab:emo-tts27}
\scriptsize
\setlength{\tabcolsep}{2.3pt}    
\renewcommand{\arraystretch}{1.15}
\begin{tabularx}{\columnwidth}{ l c *{3}{Y}}

\hline
Speaker & Model & MOS~$\uparrow$ & ESMOS~$\uparrow$ & SSMOS~$\uparrow$\\
\hline
\multirow{10}{*}{Female}
 & CosyVoice2   & $3.98 \pm 0.11$ & $3.55 \pm 0.18$ & $4.30 \pm 0.12$\\
\cline{2-5}
& CosyVoice3        & $3.92 \pm 0.12$ & $3.50 \pm 0.19$ & {\boldmath $4.36 \pm 0.11$} \\
\cline{2-5}
 & \makecell[c]{EmoInstruct-TTS\\(Dual-Path)}& {\boldmath $4.12 \pm 0.10$} & {\boldmath $3.92 \pm 0.16$} & $4.31 \pm 0.12$ \\
\cline{2-5} 
 & \makecell[c]{-w/o Emo2emb\\(Text-Only)}
   & $3.85 \pm 0.12$ & $3.32 \pm 0.20$ & $4.22 \pm 0.12$ \\
\cline{2-5} 
 & \makecell[c]{-w/o Text Instruct\\(Emo2emb-Only)}
   & $3.88 \pm 0.12$ & $3.48 \pm 0.19$ & $4.24 \pm 0.12$ \\
\hline
\multirow{10}{*}{Male}
 & CosyVoice2   & $3.90 \pm 0.12$ & $3.48 \pm 0.19$ & $4.28 \pm 0.12$ \\
\cline{2-5}
& CosyVoice3  & $3.86 \pm 0.12$ & $3.44 \pm 0.19$ & {\boldmath $4.34 \pm 0.11$} \\
\cline{2-5}
 & \makecell[c]{EmoInstruct-TTS\\(Dual-Path)} & {\boldmath $4.05 \pm 0.11$} & {\boldmath $3.78 \pm 0.17$} & $4.30 \pm 0.12$  \\
\cline{2-5} 
 & \makecell[c]{-w/o Emo2emb\\(Text-Only)}
   & $3.78 \pm 0.13$ & $3.26 \pm 0.21$ & $4.18 \pm 0.13$ \\
\cline{2-5} 
 & \makecell[c]{-w/o Text Instruct\\(Emo2emb-Only)}
   & $3.70 \pm 0.14$ & $3.20 \pm 0.22$ & $4.15 \pm 0.13$ \\
\hline
\end{tabularx}
\end{table}

\begin{table}[t]
\centering
\caption{Comparison results on 48-category emotional speech synthesis (male/female, zero-shot).
ECS and WER are reported. Best results within each speaker are highlighted in \textbf{bold}.}
\vspace{0.5em}
\label{tab:emo-tts48}
\scriptsize
\setlength{\tabcolsep}{12pt}
\renewcommand{\arraystretch}{1.15}
\begin{tabular}{ l c c }
\hline
\makecell[c]{Model} & ECS~$\uparrow$ & WER~$\downarrow$  \\
\hline
 CosyVoice2   & 0.855 & 0.0357 \\
\hline
 CosyVoice3   & 0.865 & \textbf{0.0197} \\
\hline
\makecell[l]{EmoInstruct-TTS (Dual-Path)} & \textbf{0.870} & 0.0259 \\
\hline
\makecell[l]{-w/o Emo2emb (Text-Only)} & 0.859 & 0.0329 \\
\hline
\makecell[l]{-w/o Text Instruct (Emo2emb-Only)} & 0.867 & 0.0486 \\
\hline
\end{tabular}
\end{table}

\subsection{Subjective Evaluations}
\label{ssec:Subjective}

Tables~\ref{tab:emo-tts21} and~\ref{tab:emo-tts27} report zero-shot subjective results using MOS (Mean Opinion Score), ESMOS (Emotion Similarity MOS), and SSMOS (Speaker Similarity MOS). Listening tests are conducted by 20 speech experts, with three raters per sample.

On the 21 emotion--intensity tasks, \textit{EmoInstruct-TTS (Dual-Path)} achieves the best overall MOS and ESMOS across both genders, consistently outperforming CosyVoice2 and CosyVoice3. CosyVoice3 shows competitive or slightly higher SSMOS in some cases, particularly for male speakers.

Ablation results further highlight the complementary roles of the two control paths, where removing either Emotion2embed (\textit{Text-Only}) or textual instruction (\textit{Emotion2embed-Only}) leads to noticeable performance degradation.

On the more challenging 27 fine-grained emotion tasks, EmoInstruct-TTS shows larger gains over all baselines in MOS and ESMOS, while both ablation variants degrade more substantially.

\subsection{Objective Evaluations}
\label{ssec:Objective}

Table~\ref{tab:emo-tts48} reports objective results on 48-category emotional speech synthesis using Emotion2embed Cosine Similarity (ECS) and Word Error Rate (WER). 
\textit{EmoInstruct-TTS (Dual-Path)} achieves the highest ECS (0.870), indicating more accurate emotional alignment, while \textit{CosyVoice3} obtains the lowest WER (0.0197) due to its speech-token-centric modeling. 
Overall, EmoInstruct-TTS improves emotional controllability while maintaining competitive recognition accuracy.

Ablation results show complementary roles of semantic instruction and emotion embedding. 
The \textit{Text-Only} variant degrades both ECS and WER, whereas \textit{Emotion2embed-Only} maintains relatively high ECS but suffers a large WER increase (0.0486), suggesting unstable linguistic realization without semantic guidance.




\subsection{Inference Efficiency}

Compared with the overall synthesis latency of the baseline (hundreds of milliseconds), the additional cost introduced by ICE-Flow is negligible. The MiniLM text encoder requires a single non-autoregressive forward pass ($\sim$2\,ms), and the CFM sampler (25-step Euler) adds about 3\,ms.
Overall, ICE-Flow increases the end-to-end runtime by less than 1\% for typical utterances and around 1–2\% for very short sentences.

\section{Conclusion}
\label{sec:conclusion}

This paper presents \textit{EmoInstruct-TTS}, a dual-path instruction-guided framework for controllable emotional speech synthesis. By combining natural language instructions with structured emotion embeddings, the proposed system separates semantic planning from emotion-specific acoustic control. Experiments show that EmoInstruct-TTS improves emotional controllability and speech naturalness over strong baselines, particularly for fine-grained emotion categories and intensity variations under zero-shot conditions. 
Future work will explore enabling emotional speech synthesis driven by open-ended natural language descriptions without predefined emotion categories.

\bibliographystyle{IEEEtran}
\bibliography{mybib}

\end{document}